\documentclass{article} % For LaTeX2e
\usepackage{iclr2025_conference,times}

\usepackage{amsmath,amsfonts,bm}

\def\eqref#1{equation~\ref{#1}}
\def\1{\bm{1}}

\DeclareMathAlphabet{\mathsfit}{\encodingdefault}{\sfdefault}{m}{sl}
\SetMathAlphabet{\mathsfit}{bold}{\encodingdefault}{\sfdefault}{bx}{n}

\usepackage{colortbl}
\usepackage{hyperref}
\usepackage{url}
\newcommand{\newmethod}{Multi-Layers Trainable Pooling}
\usepackage[T1]{fontenc}
\usepackage{graphicx}
\usepackage{subcaption} 
\usepackage{amsmath} 
\usepackage{amssymb}
\usepackage{booktabs}
\usepackage{array}
\usepackage{times}
\usepackage{latexsym}
\usepackage{tabularx}
\usepackage{adjustbox}
\usepackage{comment}
\usepackage{cleveref}
\usepackage{stfloats}

\title{Pooling and Attention: What are Effective Designs for LLM-based Embedding Models?}
\author{Yixuan Tang , Yi Yang \\
The Hong Kong University of Science and Technology\\
\texttt{ytangch@connect.ust.hk, imyiyang@ust.hk}
}

\iclrfinalcopy % Uncomment for camera-ready version, but NOT for submission.
\begin{document}
\maketitle
\begin{abstract}
The significant advancements of Large Language Models (LLMs) in generative tasks have led to a growing body of work exploring LLM-based embedding models. While these models, employing different pooling and attention strategies, have achieved state-of-the-art performance on public embedding benchmarks, questions still arise about what constitutes an effective design for LLM-based embedding models. However, these models are often trained on different datasets, using different LLM base models or training settings. Moreover, evaluations on public embedding benchmarks often fail to report statistical significance, making it difficult to determine which designs truly contribute to final performance. This complicates the process for practitioners seeking optimal training recipes for LLM-based embedding models.
In this study, we conduct a large-scale experiment by training a series of LLM-based embedding models using the same training data and base model but differing in their pooling and attention strategies. The results show that there is no one-size-fits-all solution: while bidirectional attention and an additional trainable pooling layer outperform in text similarity and information retrieval tasks, they do not significantly surpass simpler designs like EOS-last token pooling and default causal attention in clustering and classification tasks.
Furthermore, we propose a new pooling strategy, \newmethod, which transforms the outputs of all hidden layers, rather than just the last layer, using a cross-attention network. This method proves to be statistically superior in text similarity and retrieval tasks compared to existing pooling methods.
Overall, this paper sheds light on effective training strategies for LLM-based embedding models.
\end{abstract}

\section{Introduction}
\begin{comment}
\begin{figure*}[ht]
\centering
\includegraphics[width=0.9\textwidth]{illustration/llm2embedding.png} 
\caption{The encoding process of using LLM-based embedding model.}
\label{fig:llm2embedding}
\end{figure*}
\end{comment}

A text embedding is a high-dimensional representation that captures the semantic information of text and is crucial for many tasks, such as information retrieval and semantic textual similarity. For example, text embedding models, which convert input text into embeddings, are essential components in semantic search and retrieval-augmented generation (RAG) retrieval-augmented generation systems (RAGs) \citep{gao2023retrieval,hu2024ragrausurvey}. Companies such as OpenAI and Cohere provide embeddings as services via APIs.

Previous studies primarily employ encoder-only models, such as BERT \citep{Bert} and Sentence-BERT \citep{Sentence-BERT}, as embedding models. Recently, with the significant advancements in LLMs, the community has started to explore using LLMs as base models, fine-tuning them accordingly to serve as embedding models \citep{e5-instruct,echoEmbedding, NV-Embed}. On embedding model benchmarks such as Massive Text Embedding Benchmark (MTEB) \citep{mteb}, LLM-based embedding models show promising performance and dominate the leaderboard compared to previous encoder-only models. 

\textbf{Pooling} and \textbf{attention} are two main designs involved in converting an LLM into an embedding model. Pooling strategies are used to obtain a fixed-size dense vector representing the input sequence. For example, E5-mistral-7b-instruct \citep{e5-instruct} and SRF-Embedding-Mistral \citep{SFRAIResearch2024} use EOS-last token pooling as their pooling strategy. NV-Embed \citep{NV-Embed} uses a trainable pooling layer to obtain the final embeddings. The attention strategy constrains the direction in which tokens can attend to others. By default, an LLM is pre-trained with a causal attention mask \citep{casual_attention}, meaning that a token can only attend to preceding tokens. However, several recent works have highlighted the potential limitations of causal attention for representation learning and propose that an LLM-based embedding model should allow bidirectional attention so that every token in the sequence can attend to every other token \citep{LLM2Vec, NV-Embed}.
%For example, E5-mistral-7b-instruct uses the LLM Mistral-7b \citep{} \yi{cite} and fine-tunes it with a crafted dataset using contrastive learning \citep{e5-instruct}. %Figure \ref{fig:llm2embedding} shows the basic process of using LLM as an embedding model, and 

\begin{table*}[ht]
\centering
\resizebox{\textwidth}{!}{
\begin{tabular}{cccccc} 
    \toprule 
    \textbf{Embedding Model} & \textbf{Base Model} & \textbf{Pooling} & \textbf{Attention} & \textbf{Training Data Size} & \textbf{Score} \\
    \midrule
    e5-mistral-7b-instruct & Mistral-7B-v0.1 & EOS-Last token pool & Causal & 1.8M & 66.63 \\
    SFR-Embedding-Mistral & Mistral-7B-v0.1 & EOS-Last token pool & Causal & Not Specific & 67.56 \\
    GritLM-7B & Mistral-7B-v0.1 & Mean pool & Bidirectional  & Not Specific & 66.76 \\
    LLM2Vec-Mistral-supervised & Mistral-7B-v0.1 & Mean pool & Bidirectional & 	1.5M &
64.80 \\
    LLM2Vec-Llama-2-supervised & Llama-2-7b & Mean pool & Bidirectional & 1.5M & 64.14 \\
    NV-Embed-v1 & Mistral-7B-v0.1 & Trainable pooling layer & Bidirectional & 1.1M & 69.32 \\
    \bottomrule
\end{tabular}
}
\caption{State-of-the-art LLM-based embedding models. They vary in pooling and attention strategies. The "Score" column represents the performance on the MTEB benchmark \citep{mteb} as reported on the Hugging Face MTEB Leaderboard \protect\footnotemark}
\label{tab:exsiting_embedding_models}
\end{table*}
\footnotetext{https://huggingface.co/spaces/mteb/leaderboard}

Table \ref{tab:exsiting_embedding_models} lists state-of-the-art LLM-based embedding models. Some perform better than others. This raises the question: what makes an embedding model perform better? Is it the higher quality of the fine-tuning dataset, the greater capability of the base LLM, or the use of different pooling and attention strategies that makes the embedding model more effective? Unfortunately, most existing LLM-based embedding models are trained using different datasets with different base models, making it difficult to draw conclusions regarding the contribution of each design choice.

In this paper, we conduct large-scale experiments to empirically evaluate pooling and attention strategies for LLM-based embedding models. To ensure a fair comparison between different strategies, we fine-tune the same base LLM models (Mistral-7B and Qwen2-0.5B) using different combinations of pooling and attention strategies commonly employed in existing models. We conduct statistical testing to rigorously compare the performance of these models. Interestingly, we find that there is no one-size-fits-all solution. For example, LLMs with bidirectional attention and an additional trainable pooling layer demonstrate superior performance in semantic textual similarity (STS) and information retrieval tasks but underperform in clustering and classification tasks.

In addition to empirically testing existing pooling strategies, we also propose a new pooling strategy, \newmethod, which leverages LLM hidden states across multiple internal layers and transforms them using a trainable network. This strategy is motivated by the observation that different internal layers in LLM may encode orthogonal information that is not captured in the final layer but could be relevant for certain downstream tasks. %the hidden states of the last layer might not always be the most semantically meaningful, and that hidden states from other layers may also encode important semantic information. 
Empirical experiments show that this new pooling strategy, which pools information from multiple layers, outperforms existing methods that use only the last layer. Overall, we hope that these large-scale training experiments and the proposed pooling strategy can collectively enhance the community's efforts to improve LLM-based embedding model performance. We release the implementation of the proposed pooling method and the series of fine-tuned embedding models for replication at \url{https://github.com/yixuantt/PoolingAndAttn.git}.

% https://github.com/yixuantt/LLM_EmbeddingDesigns.git
% link

\section{Commonly Used Pooling and Attention Strategies}
In this section, we briefly review different pooling and attention strategies that are commonly used in existing LLM-based embedding models.

\subsection{Pooling Strategy}
A pooling strategy focuses on obtaining a fixed-size embedding from the LLM hidden states for an input sequence. We denote the LLM hidden states as a matrix $\mathbf{H} \in \mathbb{R}^{l \times n \times d}$, where $l$ is the hidden layer, $n$ is the sequence length, and $d$ is the hidden size.  Three commonly used pooling strategies for matrix  $\mathbf{H}$ are widely used. 

\textbf{EOS-Last Token Pooling:} $\mathbf{h}_{\text{eos}} = \mathbf{H}_{[-1, n,:]}$ 
Since the next-word prediction is often the training objective for LLM, the last token of the whole input sequence, therefore, captures all the information of the sequence. Thus, many existing works, such as OpenAI's cpt-text model \citep{cpt_text} and E5-mistral-7b-instruct \citep{e5-instruct}, append a special End-of-Sequence (EOS) token to the input and use the last layer's hidden states of the EOS token as the text embedding.  
    
\textbf{Mean Pooling:} $\mathbf{h}_{\text{mean}} = \frac{1}{n} \sum_{i=1}^{n} \mathbf{H}_{[-1, i,:]}$  Here, $\mathbf{h}_{\text{mean}}$ denotes the embedding obtained by averaging the last layer hidden states of all tokens in the sequence. Some existing works, such as GritLM-7B \citep{GritLM} and LLM2Vec \citep{LLM2Vec}, employ this pooling strategy on the LLM-based embedding model. 

\textbf{Trainable Pooling Layer:} Instead of directly using the LLM hidden states as the input embedding, NV-embed \citep{NV-Embed} pioneers a novel method using an additional trainable pooling layer to convert LLM's last layer's hidden states into a semantic latent space: $\mathbf{h}_{\text{pool}} = \mathbf{M}(\mathbf{H}_{[-1,:,:]})$, where $\mathbf{H}_{[-1,:,:]}$ is the last hidden state from LLM and  $\mathbf{M}$ is a trainable network. 

\subsection{Attention Strategy} 
LLMs are mostly pre-trained using a causal attention mask \citep{casual_attention}, a unidirectional attention mechanism that allows the current token to only attend to preceding tokens. However, several recent studies have demonstrated the limitations of unidirectional attention and have adapted the attention mask of the LLM to be bidirectional during the fine-tuning process, allowing each token to access the bidirectional context in the sequence \citep{LLM2Vec, NV-Embed,echoEmbedding}. Subsequently, we denote these two strategies as \textbf{Causal} attention and \textbf{Bidirectional} attention.

\subsection{Fine-Tuning LLMs as Embedding Models}
While existing LLM-based embedding models differ in their pooling and attention strategies, the training process is largely similar. To fine-tune an LLM as an embedding model, the contrastive learning method is often used \citep{loss, e5-instruct, NV-Embed}. In short, contrastive learning encourages the embedding of a focal example to be similar to that of a positive example while being distant from its negative example, thus enabling a base LLM to adapt to embedding-related tasks. All the works listed in Table \ref{tab:exsiting_embedding_models} use contrastive learning to fine-tune the base LLM. However, they use different training data, so the final performance may be confounded by the dataset.

\section{\newmethod: A New Pooling Strategy that Obtains Embedding From Multiple Layers}
As shown in Table \ref{tab:exsiting_embedding_models}, state-of-the-art LLM-based embedding models use pooling strategies that obtain embeddings from the last layer of the LLM's hidden states, regardless of whether they employ EOS-last token pooling, mean pooling, or trainable pooling. But can we achieve better results by using hidden states from other layers? Prior work has shown that different layers of language models, such as encoder-only BERT or decoder-only LLMs, encode different semantic information \citep{clark2019does, lastLayerBert,llmlayers}. Therefore, we hypothesize that the other layers may contain relevant information that complements the last layer. In this section, we first verify that each layer's hidden states encode distinct information and that the intermediate layers may also contain relevant information beneficial to downstream tasks.  We then propose a new pooling method, termed \textbf{\newmethod}, which consists of a trainable pooling layer that uses hidden states from all layers.

\begin{figure}[t]
\centering
\includegraphics[width=0.95\textwidth]{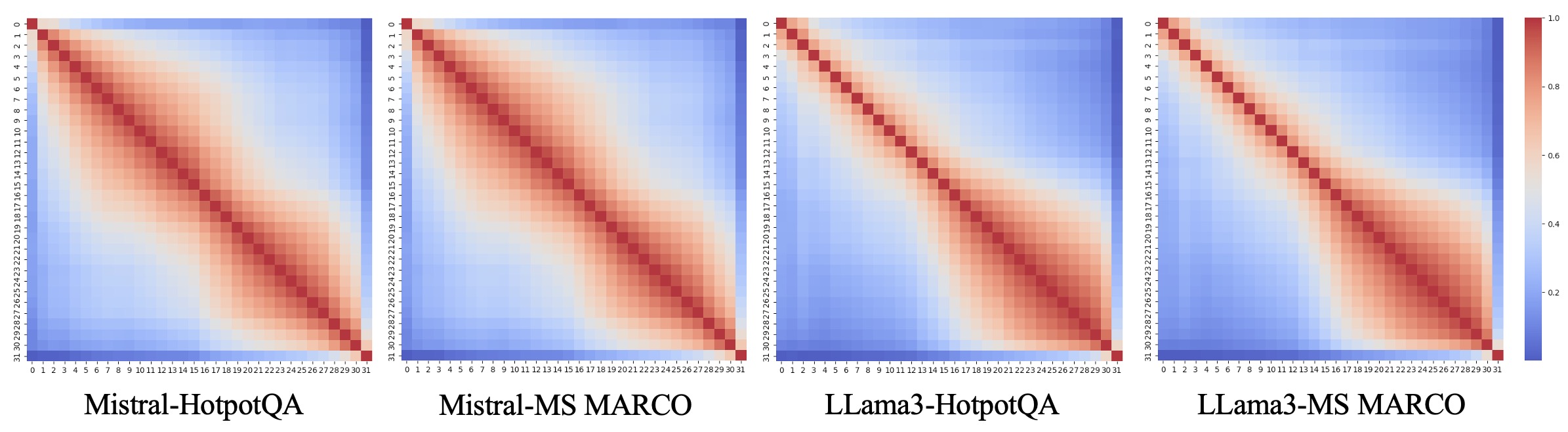} 
\caption{The correlation heatmap of EOS token hidden states across different layers. The two figures on the left are measured on Mistral-7B-v0.1 using the HotpotQA and MS MARCO datasets, while the two figures on the right are measured on Meta-Llama-3-8B. \textcolor{blue}{Areas shaded in blue} indicate low correlation, while \textcolor{red}{areas shaded in red} denote high correlation. The horizontal axis represents the layer index ranging from 0 to 31, while the vertical axis represents the layer index ranging from 31 to 0.}
\label{fig:correlation}
\end{figure}

\subsection{Last layer vs. Other layers}
We conduct two experiments to compare the hidden states of the last layer with those of other layers.

\textbf{Experiment 1: Different Hidden State Layers Encode Distinct Aspects.} In this experiment, we measure the correlation of hidden states across different layers. Specifically, we select two datasets, HotpotQA \citep{hotpotqa} and MS MARCO \citep{MSMARCO}, and append an EOS token to the input sequences. We then pass the input sequences through two LLMs, Mistral-7B-v0.1 \citep{mistral7b} and Llama3-8B \citep{llama3}, and obtain the hidden states of the EOS token from different layers. Note that neither of the base LLMs has been fine-tuned as an embedding model. For each input sequence, we measure the Spearman's correlation coefficient between the hidden states of different layers. The layer-wise correlation heatmaps are shown in Figure \ref{fig:correlation}.

The results clearly indicate that the embeddings from adjacent layers are more correlated than those from layers further apart. %The maximum window of strongly correlated adjacent layers is about 10 layers in both LLMs. 
More importantly, the findings reveal that embeddings from different layers, particularly those that are not adjacent, are vastly different and may encode distinct aspects. Although the LLMs are base models and have not been fine-tuned as embedding models, this observation suggests that the hidden states learned by different layers within LLMs are not entirely the same, indicating a variation in the information captured across the layers.

\begin{figure}[t]
\centering
\includegraphics[width=\textwidth]{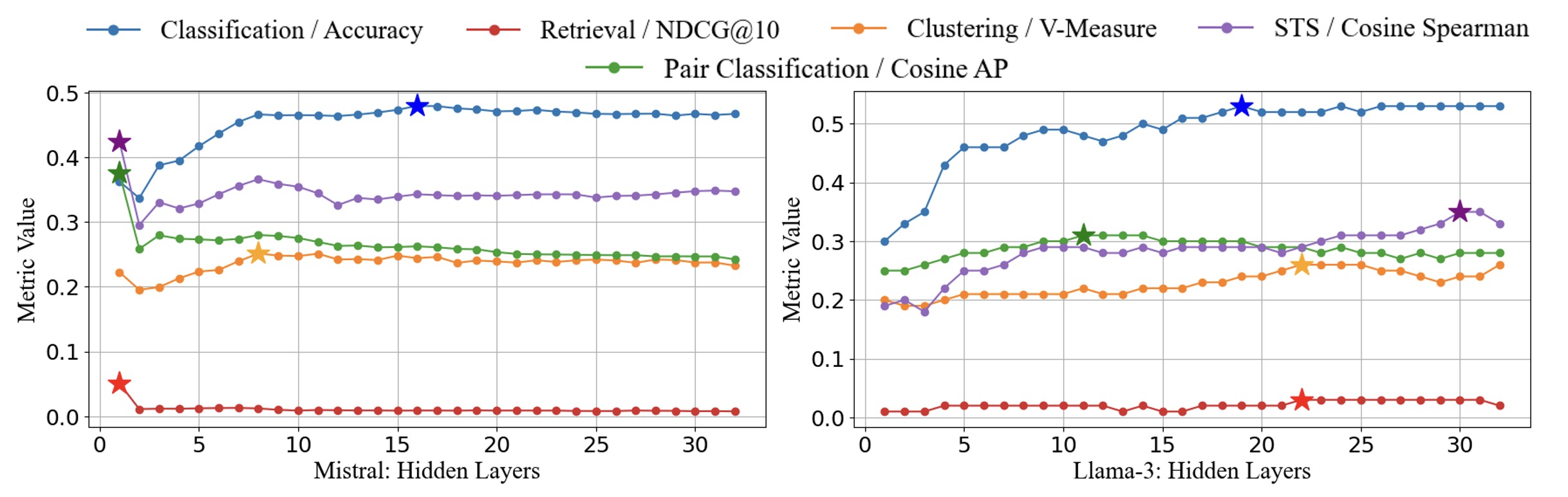} 
\caption{Performance of different hidden layers from Mistral-7B-v0.1 (left) and Llama3-8B (right) on the MTEB benchmark. The highest score is marked with a star. The X-axis represents the layer index ranging from 0 to 31, and the Y-axis represents the performance score.}
\label{fig:lastLayerTest}
\end{figure}

\textbf{Experiment 2: Other Layers' Hidden States May Also Be Useful for Downstream Tasks.} In this experiment, we assess the representation capability of the hidden states in different layers on downstream tasks. Specifically, we evaluate the downstream task performance of EOS token embedding of each layer, using the Mistral-7B-v0.1 and Llama3-8B models on the  MTEB benchmark\citep{mteb}. The results are shown in Figure \ref{fig:lastLayerTest}.

Interestingly, and perhaps surprisingly, the hidden states of the last layer do not perform the best across MTEB tasks. For Mistral-7B-v0.1, hidden states from earlier layers capture more semantic information than those from later layers. The performance gap between the last layer and the best-performing layer is 0.08 for the STS task and 0.04 for the retrieval task. For the Llama3 model, the middle layers appear to be more effective at encoding semantic meaning. 

Although the behavior of the hidden states in intermediate layers will change in fine-tuned embedding models, the key takeaway from these experiments is that the hidden states of other layers may encode information that complements that of the last layer and could be useful for downstream tasks. Thus, relying solely on the last layer's hidden state in the pooling strategy may not be optimal.

\begin{figure}[htp]
\centering
\includegraphics[width=0.96\textwidth]{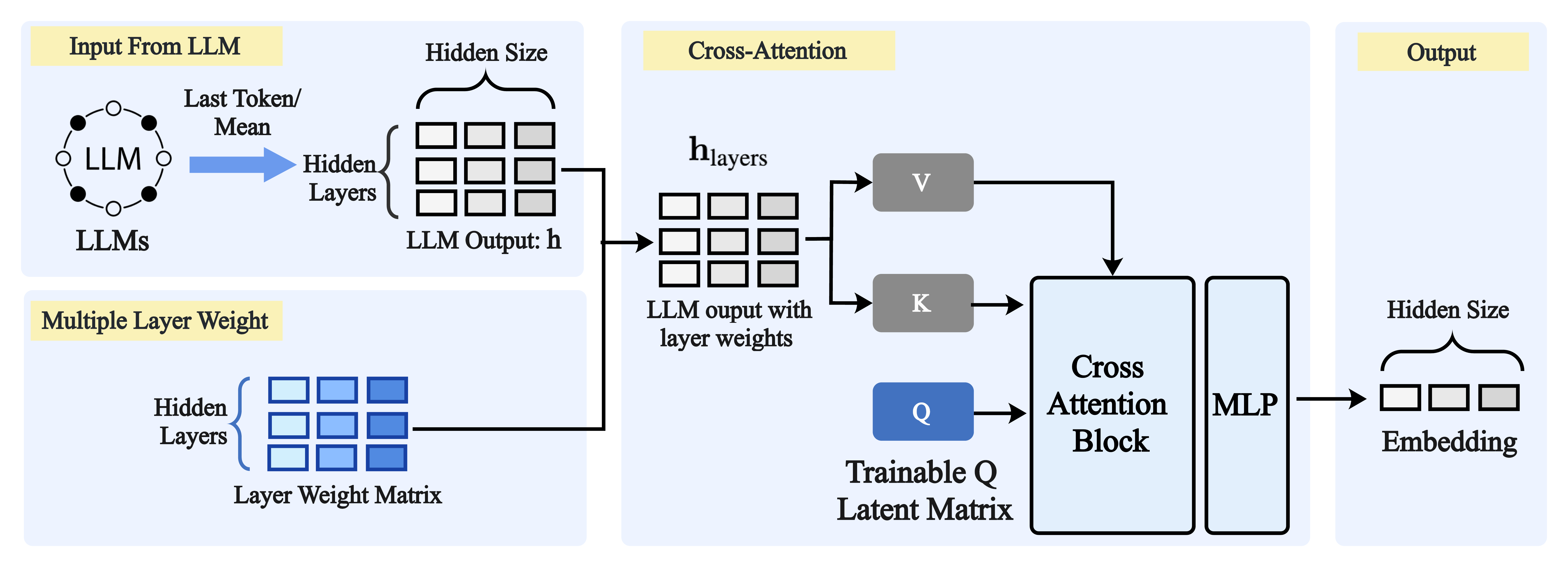} 
\caption{The proposed pooling method: \newmethod. It combines the EOS token hidden states from all layers in the LLM and transforms them into the final embedding using a cross-attention network.}
\label{fig:ourmethod}
\end{figure}

\subsection{\newmethod}
Motivated by the above findings, we propose a new pooling strategy termed \newmethod, which utilizes an additional trainable layer to capture semantic information from \textit{all layers} in an LLM. The method is shown in Figure \ref{fig:ourmethod}. Let $\mathbf{H} \in \mathbb{R}^{l \times n \times d}$ represent the LLM hidden state matrix of an input sequence, where $l$ is the number of hidden layers, $n$ is the sequence length, and $d$ is the hidden size. The high-level idea of \newmethod\ is to introduce a trainable layer that learns to pool hidden states from different layers. %As shown in Figure 2, different layers of the base LLM may encode information that is more suitable for different tasks. 
Specifically, there are three main components in this pooling method:

\textbf{Input: LLM Hidden States Across All Layers.} The first step involves selecting the LLM output $\mathbf{H}$ as the input for the pooling operation. For causal attention LLMs, we use the EOS token hidden states of all LLM layers, denoted as $\mathbf{h}_{\text{causal}} = \mathbf{H}_{[:,-1,:]}$, as the input to the subsequent trainable pooling layer. This approach is chosen because, in causal attention LLMs, earlier tokens may introduce bias, subsequently affecting the final embedding \citep{echoEmbedding}. For bidirectional attention LLMs, we consider the mean embedding across the token length dimension, expressed as $\mathbf{h}_{\text{bi-directional}} = \frac{1}{n} \sum_{i=1}^{n} \mathbf{H}_{[:, i,:]}$. The resulting vector serves as the input to the subsequent trainable pooling layer.

\textbf{Layer Weight Matrix.} Different layers may have varying importance to the final embedding depending on the task. To account for this, we introduce a trainable layer weights matrix that captures the significance of each layer. Specifically, we combine the LLM output with a trainable layer weight matrix, $\mathbf{W} \in \mathbb{R}^{l \times d}$. The combined layer matrix is computed as $\mathbf{h}_{\text{layers}} = \mathbf{h} + \mathbf{W}$, where $\mathbf{h}$ is the input (either $\mathbf{h}_{\text{causal}}$ or $\mathbf{h}_{\text{bi-directional}}$). 

\textbf{Cross Attention Matrix.} The combined layer matrix $\mathbf{h}_{\text{layers}}$ is then multiplied by the parameter matrices $W_K$ and $W_V$  and produce the key matrix $\mathbf{K}$ and value matrix $\mathbf{V}$ accordingly.  The $\mathbf{K}$ and $\mathbf{V}$ matrix is then combined with a trainable query matrix $\mathbf{Q}$ $\in \mathbb{R}^{r \times d'}$, with $d'$ being the inner dimension of the cross-attention block and $r$ being the number of latent dimensions which is the same with LLM hidden dimension. Note that the $\mathbf{K}$ and $\mathbf{V}$ are both derived from the linear transformation of input $\mathbf{h}_{\text{layers}}$, while $\mathbf{Q}$ is a trainable parameter matrix. This cross-attention network is similar to the method used in Flamingo \citep{Flamingo}, which maps varying-size video frames to fixed-size visual embeddings, similar to obtaining embeddings from multiple layers.

The cross-attention network computes attention between fixed, trainable queries and keys/values derived from the input to capture and encode the most relevant information from $\mathbf{h}_{\text{layers}}$ into a semantic latent space. The output of this cross-attention network is then passed through a multi-layer perceptron (MLP) to produce the final output embedding. Using a trainable query matrix ($\mathbf{Q}$) instead of deriving it directly from the input data ($\mathbf{h}_{\text{layers}}$) allows the cross-attention mechanism to more effectively filter semantic information from multi-layer hidden states through the trainable $\mathbf{Q}$.

% We denote the combined input $\mathbf{h}_{\text{layers}}$ which is from LLM hidden states and layer weight matrix as the matrix $\mathbf{K} = \mathbf{V} \in \mathbb{R}^{r \times d} $. The dimension $d$ is consistent with the LLM hidden dimension. We employ a trainable matrix $\mathbf{Q} \in \mathbb{R}^{r \times d}$ for cross-attention operations to capture the most relevant information from the input and encode it into a latent space. The dimension $r$ is the number of latent in the dictionary. The output of the cross-attention operation is then passed through MLP layers to obtain the final input embedding. %The whole process can be illustrated as: \yi{need a formula?}$$O = \text{softmax}(QK^T) V$$

Compared to NV-embed \citep{NV-Embed}, which pioneered the use of a trainable pooling layer, our proposed pooling method introduces three key innovations. First, we utilize hidden states from all layers rather than just the last layer. Second, we include a trainable layer weight matrix to account for the varying importance of different layers across tasks. Third, while NV-embed transforms the last layer's hidden states to the query matrix $\mathbf{Q}$, our method combines the hidden state matrix with the layer weight matrix and transforms it into key matrix $\mathbf{K}$ and value matrix $\mathbf{V}$.

\section{Pooling and Attention Experiments}
%To analyze the best practice for it, we strategically select and configure five distinct settings  %These configurations are designed to systematically assess the impact of different pooling  and attention methods on model performance, addressing subtle distinctions revealed by prior research and advancing our understanding of optimal embedding strategies.
As shown in Table 1, state-of-the-art embedding models are often trained using different pooling and attention strategies. However, two factors obscure our understanding of the most effective designs for LLM-based embedding models. First, they are trained using different datasets, which is a key confounding factor that significantly influences final performance on the MTEB benchmarks. Second, they primarily report results without statistical testing, making it unclear whether the observed performance improvements are statistically meaningful. %With various pooling and attention strategies available for converting an LLM into an embedding model, determining the optimal combination remains an open question. 
To address these, we aim to empirically assess the effectiveness of different pooling and attention strategies through a fair comparison using the same dataset and training protocol.

\subsection{Pooling and Attention Combinations} 
In the experiment, we consider the following five design combinations: 

\noindent\textbf{Model 1:} EOS-Last Token Pooling + Causal Attention

\noindent\textbf{Model 2:} Last-Layer Trainable Pooling + Causal Attention

\noindent\textbf{Model 3:} \newmethod + Causal Attention

\noindent\textbf{Model 4:} Last-Layer Trainable Pooling + Bidirectional Attention

\noindent\textbf{Model 5:} \newmethod + Bidirectional attention

These five combinations allow us to conduct pairwise comparisons between different pooling and attention strategies by controlling for other potential confounding factors. For example, by comparing Model 1, Model 2, and Model 3, we can assess the effectiveness of different pooling strategies under the same attention method. Similarly, by comparing Model 3 and Model 5, we can evaluate how different attention methods affect embedding performance when using a multi-layer trainable pooling strategy.

%We implement the last-layer trainable pooling, similar to that used in NV-Embed \citep{NV-Embed}. It consists of a cross-attention block and an MLP layer to transform the hidden states of the last layer to an embedding space. %To demonstrate the performance of using hidden states from multiple layers, we keep the parameters and modules of the Last-Layer Trainable Pooling consistent with the \newmethod, except for a trainable Layer Weights component. This means that the input of the Last-Layer Trainable Pooling is the last hidden state, sent directly to the $K$ and $V$ matrices without the addition of trainable layer weights. \yi{add a footnote explain the little difference between here and nv-embed}

\textbf{Why Mean Pooling Is Not Considered in Our Setting?} First, prior work has demonstrated that employing mean pooling in a causal attention LLM-based embedding model introduces a bias towards the earlier tokens \citep{echoEmbedding}, leading to poor performance \citep{e5-instruct,LLM2Vec}. Second, for LLMs that use bidirectional attention, NV-embed \citep{NV-Embed} has shown that an additional trainable pooling layer can outperform mean pooling. Thus, mean pooling does not appear to be a viable choice for either attention strategy. To keep our experiment manageable in scope, we have therefore excluded mean pooling from consideration.

\textbf{Why EOS-Last Token Pooling + Bidirectional Attention Is Not Considered in Our Setting?} Intuitively, when bidirectional attention is used, the EOS-Last token is no longer meaningful as the input embedding. In fact, existing LLM-based embedding models that use bidirectional attention typically employ pooling techniques such as mean pooling or trainable pooling layers \citep{NV-Embed, LLM2Vec, GritLM}. Therefore, in our experiment, we do not include this combination of pooling and attention.

\subsection{Experimental Details}
\textbf{Base LLM.} We use the Mistral-7B-v0.1 model as the base LLM. We choose Mistral-7B because it is widely regarded as one of the best open-source LLM models for embeddings and is commonly used in state-of-the-art embedding models, as shown in Table 1.

\textbf{Training data.} We use publicly available datasets that are commonly utilized for embedding model fine-tuning to train Model 1 to Model 5. Since the models are trained on the same dataset but differ in either pooling or attention strategy, this approach allows us to isolate the impact of these strategies and ensures a fair comparison between them. The size of the training dataset is 1.4 million, which is about the same scale as existing works illustrated in Table \ref{tab:exsiting_embedding_models}. 
An additional EOS token is appended to the end of each training example. 
Following e5-mistral-7b-instruct \citep{e5-instruct}, we also include instructions in the query to describe the task.
Table \ref{tab:training_dataset} in the Appendix lists the training datasets and associated instructions used in this study. 

\textbf{Contrastive Learning.} For contrastive learning, we follow the standard training pipeline that each query is paired with one positive and one hard negative example \citep{e5-instruct}. The positive example is provided by the datasets, while the hard negative example is mined by a trained SentenceTransformer \footnote{https://huggingface.co/sentence-transformers/all-MiniLM-L6-v2}. During training, we utilize in-batch negatives, where the negative examples for a given query are sourced from the other queries within the same batch.

\textbf{Training Setting.} 
We use Low-Rank Adaptation (LoRA) \citep{lora} with a LoRA rank of 16 to finetune the model for downstream embedding tasks using contrastive learning loss. The learning rate is 1e-5, and the training batch size is 2,048. The max training step is 1,000, which is aligned with existing works \citep{e5-instruct,LLM2Vec}. 

\textbf{Trainable Pooling Layers Setting.}
The Last-Layer Trainable Pooling employs query and cross-attention dimensions consistent with the hidden size of the Large Language Model (LLM), which is 4,096 for Mistral-7B. The multi-head cross-attention mechanism utilizes 32 heads, each containing 2,048 channels. This setting aims to reproduce a module similar to NV-embed \citep{NV-Embed}. The proposed \newmethod\ shares the same basic module parameters as the Last-Layer Trainable Pooling, except for the trainable layer weight matrix. Specifically, the trainable layer weight is $\mathbf{W} \in \mathbb{R}^{32 \times 4096}$.

\textbf{Evaluation.} 
We evaluate all five fine-tuned models on the  MTEB Benchmark \citep{mteb} encompassing 15 retrieval datasets, 4 reranking datasets, 12 classification datasets, 11 clustering datasets, 3 pair classification datasets, 10 semantic textual similarity datasets, and 1 summarization dataset. Table \ref{tab:eval_dataset} in the Appendix lists all the evaluation tasks and instructions.  

\textbf{Wilcoxon Signed Rank Test.} While the MTEB benchmark is commonly used in prior LLM-based embedding models, it is unfortunate that statistical significance is not commonly reported. As a result, it remains unclear whether the improvement of a specific design, such as bidirectional attention, is statistically meaningful. The MTEB benchmark comprises seven tasks, with each task containing a different number of datasets. To ensure statistical rigor and accurately assess model performance on each task, we conduct a Wilcoxon Signed Rank Test within each task to determine the statistical significance of the experimental results. Tasks with four or fewer datasets—including reranking, pair classification, and summarization—are excluded from this test due to the extremely small sample size, which limits the statistical power of the analysis. Therefore, we employ the Wilcoxon Signed Rank Test on the Retrieval, STS, Classification, and Clustering tasks. \textbf{We consider the comparison to be significant when the p-value is less than 0.05.}.

\section{Empirical Analysis}
After fine-tuning the Mistral-7B with various configurations (Model 1 $\sim$ Model 5) and testing on the MTEB benchmark, we provide an empirical analysis of the effectiveness of different pooling and attention strategies.

% \subsection{What is the Optimal Pooling Strategy?}
% \begin{table}[ht]
% \resizebox{\textwidth}{!}{
% \centering
% \begin{tabular}{cccccccc}
% \toprule
% \textbf{Pooling Strategy }& \textbf{STS} & \textbf{Clas.} & \textbf{Retr.} & \textbf{Clus.} &\textbf{ Rera. }& \textbf{P. Cls.} & \textbf{Summ.} \\ \midrule
% \multicolumn{8}{c}{\textbf{Casual Attention}} \\
% \midrule 
% \rowcolor{gray!20} 
% EOS-Last Token Pooling & 0.8302 & 0.7244 & 0.5394 & 0.4503 & 0.5737 & 0.8605 & 0.3240 \\
% Last-Layer Trainable Pooling & \textbf{+0.0129*} & -0.0035 & +0.0102 & -0.0076 & -0.0017 & +0.0034 & -0.0143 \\ 

% %\hline
% %\rowcolor{gray!20} 
% %EOS Token Pooling & 0.8302 & 0.7244 & 0.5394 & 0.4503 & 0.5737 & 0.8605 & 0.3240 & 0.6146\\
% \newmethod & \textbf{+0.0118*} & -0.0033 & +0.0135 & -0.0017 & +0.005 & +0.0022 & -0.0244 \\

% \hline
% \rowcolor{gray!20} 
% Last-Layer Trainable Pooling & 0.8431  & 0.7209 & 0.5496 & 0.4427 & 0.5720 & 0.8639 & 0.3097 \\ 
% \newmethod & -0.0011 & +0.0002 & +0.0033 & +0.0059 & +0.0067 & -0.0012 & -0.0101 \\

% \midrule
% \multicolumn{8}{c}{\textbf{Bi-directional Attention}} \\
% \midrule 
% \rowcolor{gray!20} 
% Last-Layer Trainable Pooling & 0.8397 & 0.6761 & 0.5607 & 0.4010 & 0.5829 & 0.8707 & 0.2979\\ 
% \newmethod & \textbf{+0.0071*} & \textbf{+0.034*} & \textbf{+0.0013*} & \textbf{+0.0247*} & -0.0017 & +0.0039 & -0.0033 \\
% \bottomrule
% \end{tabular}
% }

\subsection{What is the Optimal Pooling Strategy?}
\begin{table}[htp]
\centering
 \resizebox{0.95\textwidth}{!}{
\begin{tabular}{cccccc}
\toprule
\textbf{Combination }& \textbf{Pooling}& \textbf{STS} & \textbf{Clas.} & \textbf{Retr.} & \textbf{Clus.} \\ \midrule
\multicolumn{6}{c}{\textbf{Casual Attention}} \\
\midrule 
\rowcolor{gray!20} 
Model 1 & EOS-Last Token Pooling & 0.8302 & 0.7244 & 0.5394 & 0.4503 \\
Model 2 & Last-Layer Trainable Pooling & \textbf{+0.0129*} & -0.0035 & +0.0102 & -0.0076  \\ 
Model 3 & \newmethod & \textbf{+0.0118*} & -0.0033 & +0.0135 & -0.0017  \\

\hline
\rowcolor{gray!20} 
Model 2 & Last-Layer Trainable Pooling & 0.8431  & 0.7209 & 0.5496 & 0.4427  \\ 
Model 3 &  \newmethod & -0.0011 & +0.0002 & +0.0033 & +0.0059  \\

\midrule
\multicolumn{6}{c}{\textbf{Bi-directional Attention}} \\
\midrule 
\rowcolor{gray!20} 
Model 4 &  Last-Layer Trainable Pooling & 0.8397 & 0.6761 & 0.5607 & 0.4010 \\ 
Model 5 & \newmethod & \textbf{+0.0071*} & \textbf{+0.034*} & \textbf{+0.0013*} & \textbf{+0.0247*} \\
\bottomrule
\end{tabular}
}
\caption{Comparison of pooling strategies on the MTEB benchmark. The score represents the average score across datasets within each task. The row in \colorbox{gray!20}{gray} is the baseline for comparison, and the pairwise significant results are marked with asterisks* with a p-value less than 0.05. }
\label{tab:pooling_strategy} 
\end{table}

In this section, we compare EOS token pooling, Last-Layer Trainable Pooling, and proposed \newmethod\. The Last-Layer Trainable Pooling is similar to \newmethod\, but the input contains only the last hidden states and without trainable layer weights. The results are illustrated in the Table \ref{tab:pooling_strategy}.

\textbf{Finding (1): An additional trainable pooling layer is preferable only for STS tasks, but not other tasks, when causal attention is used.} For the STS task, the performance of EOS-Last token pooling is statistically lower than using a trainable pooling layer (regardless of whether using Last-Layer Trainable Pooling or Multi-Layers Trainable Pooling). As shown in Table \ref{tab:pooling_strategy}, the Last-Layer Trainable Pooling and \newmethod\ both achieve significant improvement (+0.0129 and +0.018 respectively) compared to EOS token pooling. However, both trainable pooling methods fail to pass the significance test for the classification, retrieval, and clustering tasks.%This suggests that a trainable pooling layer, which captures a more comprehensive representation of input embedding, is better for tasks requiring high semantic understanding. 

\textbf{Finding (2): Using Multi-Layers trainable pooling is more effective than Last-Layer trainable pooling when bidirectional attention is used. } As shown in the bottom rows of Table 2, Multi-Layers Trainable Pooling significantly outperforms Last-Layer Trainable Pooling across all four tasks (STS, Classification, Retrieval, and Clustering).  However, no significant results are observed in the causal attention LLM setting across four tasks using the Wilcoxon Signed Rank Test, suggesting that the benefits of multi-layer pooling are context-dependent.

\subsection{What is the Optimal Attention Strategy?}
\begin{table}[ht]
\centering
 \resizebox{0.98\textwidth}{!}{
\begin{tabular}{ccccccc}
\toprule
\textbf{Combination }& \textbf{Attention}& \textbf{STS} & \textbf{Clas.} & \textbf{Retr.} & \textbf{Clus.} \\ \midrule
\multicolumn{6}{c}{\textbf{Last-Layer Trainable Pooling}} \\
\midrule 
\rowcolor{gray!20} 
Model 2 & Casual Attention & 0.8431  & 0.7209 & 0.5496 & 0.4427 \\
Model 4 &Bi-directional Attention & -0.0034 & -0.0448 & \textbf{+0.0111*} & \textbf{-0.0417*}  \\
\midrule 
\multicolumn{6}{c}{\textbf{\newmethod}} \\
\midrule 
\rowcolor{gray!20} 
Model 3 & Casual Attention & 0.8420 & 0.7211 & 0.5529 & 0.4486 \\
Model 5 & Bi-directional Attention  & +0.0048 & -0.011 & \textbf{+0.0091*} & \textbf{-0.0229*} \\
\bottomrule
\end{tabular}
}
\caption{Comparison of attention strategies on the MTEB benchmark. The score represents the average score across datasets within each task. The row in \colorbox{gray!20}{gray} is the baseline for comparison, and the pairwise significant results are marked with asterisks*. }
\label{tab:attention_mask} 
\end{table}

\textbf{Finding (3): Bi-directional attention is better at retrieval task but worse at clustering task.} 
The data from Table \ref{tab:attention_mask} demonstrates that bi-directional attention masks consistently improve performance in the retrieval tasks, regardless of the pooling strategy employed, although the absolute improvement on the retrieval task is trivial. In contrast, the same configuration leads to diminished performance in the clustering tasks, as indicated by the negative deltas in scores (-0.0417 for Last-Layer Trainable Pooling and -0.0229 for \newmethod). These divergences suggest that the bidirectional attention strategy enhances the model's capacity to consider the context from both directions, proving beneficial for retrieving relevant information. However, this increased context may also introduce noise, which can hinder effective clustering.
% Table \ref{tab:attention_mask} shows that a bi-directional attention mask LLM is better at retrieval tasks but poor at clustering. This finding remains consistent with different external pooling layers. Thus, the choice of attention masks depends on the downstream task.

\subsection{What is the Optimal Pooling and Attention Design?}
The performance of Model 1 $\sim$ Model 5 is presented in  Table \ref{tab:recipe}.

\begin{table*}[ht]
\resizebox{\textwidth}{!}{
\centering
\begin{tabular}{ccccccc}
\toprule
\textbf{Combination }&  \textbf{Pooling}& \textbf{Attention} &\textbf{STS} & \textbf{Clas.} & \textbf{Retr.} & \textbf{Clus.} \\ \midrule
\rowcolor{gray!20} 
Model 1 & EOS-Last Token Pooling & Casual& 0.8302 & 0.7244 & 0.5394 & 0.4503  \\
Model 2 & Last-Layer Trainable Pooling & Casual & \textbf{+0.0129*} & -0.0035 & +0.0102 & -0.0076 \\ 
Model 3 &  \newmethod & Casual & \textbf{+0.0118*} & -0.0033 & +0.0135 & -0.0017  \\
Model 4 &Last-Layer Trainable Pooling & Bi-directional & \textbf{+0.0095*} & -0.0483 & +0.0213 & \textbf{-0.0493*} \\
Model 5 &\newmethod & Bi-directional & \textbf{+0.0166*} & \textbf{-0.0143*} & \textbf{+0.0226*} & \textbf{-0.0246*}  \\

\bottomrule
\end{tabular}
}
\caption{Comparison of different pooling and attention combinations on MTEB benchmark. \textbf{Mistral-7B-v0.1} is the base LLM. The row in \colorbox{gray!20}{gray} is the baseline for comparison, and the pairwise significant results are marked with asterisks*.}
\label{tab:recipe} 
\end{table*}

\textbf{Finding (4): There is no one-size-fits-all winner.} 
The results in Table \ref{tab:recipe} demonstrate the varying efficacy of different pooling and attention strategies across multiple tasks. For the STS and retrieval tasks, Multi-Layers Trainable Pooling + Bidirectional (Model 5) significantly and substantially outperforms the other models. For example, Model 5 achieves a 4.2\% improvement (+0.0226) over Model 1, which is a standard setting in training LLM-based embedding models, on the retrieval task. However, the Model 5 configuration is less effective for the classification and clustering task, where a more directed, causal attention strategy performs significantly higher.  Therefore, there is no one-size-fits-all solution, and the effectiveness of pooling and attention strategies appears to be task-dependent. That being said, we believe that the superiority in STS and retrieval tasks suggests the Multi-Layers Trainable Pooling + Bidirectional (Model 5) might be a more viable choice for practitioners, given that LLM-based embeddings are ``essential building blocks for semantic search and retrieval-augmented generation (RAG), which is the predominant approach for domain-specific or company-specific chatbots and other AI application''\footnote{VoyageAI. (September 4, 2024 version), https://docs.voyageai.com/docs/introduction}.

\section{Robustness Check: Qwen as the base LLM}
To confirm the robustness of our analysis efficiently, we further fine-tune Model 1 $\sim$ Model 5 based on Qwen2-0.5B \citep{qwen2} using the same training data and evaluate their performance on the MTEB benchmark. The results are presented in Table \ref{tab:ablation}. The reason for choosing this smaller base model is that an embedding model based on a smaller LLM might be more suitable and inference-efficient in resource-constrained situations, and training a smaller LLM-based embedding model is practically relevant \citep{qwen2GTE}. The findings are largely consistent.

First, the overall performance across the four tasks is lower than that reported in Table 4, indicating a significant capacity gap between the base LLMs, Mistral-7B and Qwen2-0.5B. Second, similarly Multi-Layers Trainable Pooling + Bidirectional (Model 5) significantly outperforms the other models on the STS but performs significantly worse on the classification task. This result aligns with Finding 2 and Finding 3, demonstrating that Multi-Layers Trainable Pooling and Bidirectional attention are capable of encoding more contextual information, which is advantageous for retrieval and STS tasks. However, for classification tasks, such as news article classification, the directionality of a text sequence may be less critical, as global semantics play a more crucial role in classification. Third, except for the STS task, the simple EOS-Last token pooling with causal attention (Model 1) performs on par with the trainable pooling methods and bidirectional attention. This suggests that for embedding models based on a smaller LLM, employing more complex designs in pooling and attention strategies does not yield meaningful gains. 
Moreover, the mixed results underscore the complexity of benchmarking an embedding model and advocate for researchers to conduct statistical significance tests when reporting results.
% Besides, though all alternatives surpass the baseline, bi-directional attention with \newmethod\ still performs the best. Those finds confirm the insights (3) and (4).

\begin{table*}[ht]
\resizebox{\textwidth}{!}{
\centering
\begin{tabular}{ccccccc}
\toprule
\textbf{Combination }&  \textbf{Pooling  }& \textbf{Attention } &\textbf{STS} & \textbf{Clas.} & \textbf{Retr.} & \textbf{Clus.} \\ \midrule
\rowcolor{gray!20} 
Model 1 & EOS Token Pooling & Casual & 0.7765 & 0.6903 &0.3867 & 0.3885 \\
Model 2 & Last-Layer Trainable Pooling & Casual &\textbf{ +0.0250*} & -0.0183 & -0.0280 & -0.0078  \\ 
Model 3 & \newmethod & Casual &\textbf{ +0.0268*} & -0.0347 & -0.0084 & -0.0035  \\
Model 4 & Last-Layer Trainable Pooling & Bi-directional & \textbf{+0.0234*} & \textbf{-0.0333*} & -0.0013 & +0.0103  \\
Model 5 &\newmethod & Bi-directional & \textbf{+0.0372*} & \textbf{-0.0393*} & +0.0003 & +0.0019  \\
\bottomrule
\end{tabular}
}
\caption{Comparison of different pooling and attention combinations on MTEB benchmark. \textbf{Qwen2-0.5B} is the base LLM. The row in \colorbox{gray!20}{gray} is the baseline for comparison, and the pairwise significant results are marked with asterisks*.}
\label{tab:ablation} 
\end{table*}

\section{Related Works}
\subsection{Encoder-based Embedding Models}
Text embedding has evolved with the Transformer architecture \citep{attentionneed}. Encoder models, particularly BERT \citep{Bert} and the T5 Encoder \citep{T5-Encoder}, have been widely used in tasks like text similarity by capturing sentence-level semantics. Building upon this foundation, Sentence-BERT \citep{Sentence-BERT} uses Siamese networks for fixed-size embeddings to efficiently retrieve semantically similar sentences. Furthermore, the INSTRUCTOR model \citep{hkuInstructer} leverages the T5 Encoder to incorporate various instructional prompts, allowing it to adapt to a wide range of downstream tasks. The BGE-M3 model \citep{bge-m3}, based on XLM-RoBERTa \citep{XLMRobertaModel}, integrates dense and sparse retrieval to support multi-granularity in the retrieval process. 

% Word embedding, a fundamental task in natural language processing (NLP), has seen advancements following the development of the Transformer architecture \citep{attentionneed}. Encoder-based models have become predominant for embedding tasks, such as text similarity, with the hidden states from models like BERT \citep{Bert} and the T5 Encoder \citep{T5-Encoder} being utilized to capture sentence-level semantics. Building upon this foundation, Sentence-BERT \citep{Sentence-BERT} employs Siamese BERT-Networks to generate fixed-size sentence embeddings. This approach enables efficient comparison and retrieval of sentences that are semantically similar. Furthermore, the INSTRUCTOR model \citep{hkuInstructer} leverages the T5 Encoder to incorporate various instructional prompts, allowing it to adapt to a wide range of downstream tasks. The BGE-M3 \citep{bge-m3}, which is based on XLM-RoBERTa \citep{XLMRobertaModel}, supports both embedding and sparse retrieval. 

\subsection{LLM-based Embedding Models}
The success of LLMs in text generation tasks has sparked increasing interest in exploring LLM-based embedding models. RepLLama \citep{RepLLAMa} pioneered this promising direction by finetuning an LLM as a dense retriever, demonstrating the potential of LLMs in embedding tasks. Building upon this foundation, subsequent research has explored various techniques to enhance LLM-based embedding models. E5-mistral-7b-instruct \citep{e5-instruct} investigated the usage of synthetic data in the training process.  Recognizing the significance of capturing bidirectional context in embedding tasks, GritLM \citep{GritLM} and LLM2Vec \citep{LLM2Vec} employ bidirectional attention mechanisms in their LLM architectures. By attending to both past and future tokens, these models can generate more contextually informed embeddings, potentially leading to improved performance in tasks such as text similarity and retrieval. Furthermore, NV-Embed \citep{NV-Embed} introduces a novel latent attention layer to obtain pooled embeddings for a sequence of tokens. These advancements showcase the ongoing efforts to harness the power of LLMs for embedding tasks. However, existing LLM-based embedding models are often trained on different datasets, leading to mixed conclusions regarding the effectiveness of pooling and attention strategies. Our work aims to empirically evaluate and deepen our understanding of the training design choices for LLM-based embeddings.

% \subsection{Embedding Model Evaluation}
% To demonstrate and rank the embedding models, there are some widely used embedding benchmarks. The early one is Benchmarking-IR (BEIR) \citep{beir}, which is mainly focused on the information retrieval tasks. Besides, The Text Embedding Benchmark (MTEB) benchmark \citep{mteb}, which integrates a large number of existing datasets across 7 tasks, including classification, clustering, pair-classification, re-ranking, retrieval, Semantic Textual Similarity (STS), and Summarization serves as a comprehensive testbed for different embedding models. Based on it, there are also some variations, such as the Chinese Massive Text Embedding Benchmark (C-MTEB) \citep{bge_embedding}, which is similar to the MTEB benchmark but derived from Chinese text and MTEB-French \citep{MTEB-French}, which is derived from French corpus. This work is also based on the MTEB benchmark leveraging its various datasets and tasks to provide a robust conclusion about the effective design choice of the LLM-based embedding model. 

\section{Conclusion}
In this study, we investigate LLM-based embedding models, focusing on two key design elements: pooling and attention. We conduct a large-scale experiment by fine-tuning five LLM-based embedding models on the same training data using different pooling and attention strategies. Our findings highlight that fine-tuning LLMs with bidirectional attention and an additional trainable pooling layer demonstrates superior performance in semantic textual similarity and information retrieval tasks, but underperforms in clustering and classification tasks. Furthermore, we introduce a new and effective pooling method, \newmethod, which leverages all layers rather than just the last hidden layer to capture broader and potentially more relevant semantic information. %The empirical results from our experiment confirm that \newmethod consistently outperforms existing pooling methods that rely solely on the last layer of hidden states in semantic textual similarity and information retrieval tasks. However, \newmethod's performance is on par with simple EOS-Last token pooling and default causal attention in clustering and classification tasks, indicating that performance is task-dependent. 
We hope this work sheds light on training LLM-based embedding models.

%\section*{Limitations}
%This study conducts comprehensive control experiments to investigate the model architecture. However, the standards for high-quality data suitable for training large language model (LLM) based embedding models remain unexplored. The importance of training data is demonstrated in the e5-mistral-7b-instruct \citep{e5-instruct}, highlighting the need for future research to establish best practices for curating datasets optimized for LLM-based embedding models.
%The importance of training data is demonstrated in the E5-mistral-7b-instruc \citep{e5-instruct}, highlighting the need for future research to establish best practices for curating datasets optimized for LLM-based embedding models.


\bibliography{custom}

\begin{thebibliography}{46}
\expandafter\ifx\csname natexlab\endcsname\relax\def\natexlab#1{#1}\fi

\bibitem[{Alayrac et~al.(2022)Alayrac, Donahue, Luc, Miech, Barr, Hasson, Lenc, Mensch, Millican, Reynolds et~al.}]{Flamingo}
Jean-Baptiste Alayrac, Jeff Donahue, Pauline Luc, Antoine Miech, Iain Barr, Yana Hasson, Karel Lenc, Arthur Mensch, Katherine Millican, Malcolm Reynolds, et~al. 2022.
\newblock Flamingo: a visual language model for few-shot learning.
\newblock \emph{Advances in neural information processing systems}, 35:23716--23736.

\bibitem[{Bajaj et~al.(2016)Bajaj, Campos, Craswell, Deng, Gao, Liu, Majumder, McNamara, Mitra, Nguyen et~al.}]{MSMARCO}
Payal Bajaj, Daniel Campos, Nick Craswell, Li~Deng, Jianfeng Gao, Xiaodong Liu, Rangan Majumder, Andrew McNamara, Bhaskar Mitra, Tri Nguyen, et~al. 2016.
\newblock Ms marco: A human generated machine reading comprehension dataset.
\newblock \emph{arXiv preprint arXiv:1611.09268}.

\bibitem[{BehnamGhader et~al.(2024)BehnamGhader, Adlakha, Mosbach, Bahdanau, Chapados, and Reddy}]{LLM2Vec}
Parishad BehnamGhader, Vaibhav Adlakha, Marius Mosbach, Dzmitry Bahdanau, Nicolas Chapados, and Siva Reddy. 2024.
\newblock Llm2vec: Large language models are secretly powerful text encoders.
\newblock \emph{arXiv preprint arXiv:2404.05961}.

\bibitem[{Bowman et~al.(2015)Bowman, Angeli, Potts, and Manning}]{allnil}
Samuel~R. Bowman, Gabor Angeli, Christopher Potts, and Christopher~D. Manning. 2015.
\newblock \href {https://doi.org/10.18653/v1/D15-1075} {A large annotated corpus for learning natural language inference}.
\newblock In \emph{Proceedings of the 2015 Conference on Empirical Methods in Natural Language Processing}, pages 632--642, Lisbon, Portugal. Association for Computational Linguistics.

\bibitem[{Cer et~al.(2017)Cer, Diab, Agirre, Lopez-Gazpio, and Specia}]{cer-etal-2017-semeval}
Daniel Cer, Mona Diab, Eneko Agirre, I{\~n}igo Lopez-Gazpio, and Lucia Specia. 2017.
\newblock \href {https://doi.org/10.18653/v1/S17-2001} {{S}em{E}val-2017 task 1: Semantic textual similarity multilingual and crosslingual focused evaluation}.
\newblock In \emph{Proceedings of the 11th International Workshop on Semantic Evaluation ({S}em{E}val-2017)}, pages 1--14, Vancouver, Canada. Association for Computational Linguistics.

\bibitem[{Chen et~al.(2024)Chen, Xiao, Zhang, Luo, Lian, and Liu}]{bge-m3}
Jianlyu Chen, Shitao Xiao, Peitian Zhang, Kun Luo, Defu Lian, and Zheng Liu. 2024.
\newblock \href {https://aclanthology.org/2024.findings-acl.137} {{M}3-embedding: Multi-linguality, multi-functionality, multi-granularity text embeddings through self-knowledge distillation}.
\newblock In \emph{Findings of the Association for Computational Linguistics ACL 2024}, pages 2318--2335, Bangkok, Thailand and virtual meeting. Association for Computational Linguistics.

\bibitem[{Clark(2019)}]{clark2019does}
Kevin Clark. 2019.
\newblock What does bert look at? an analysis of bert’s attention.
\newblock \emph{arXiv preprint arXiv:1906.04341}.

\bibitem[{Conneau et~al.(2020)Conneau, Khandelwal, Goyal, Chaudhary, Wenzek, Guzm{\'a}n, Grave, Ott, Zettlemoyer, and Stoyanov}]{XLMRobertaModel}
Alexis Conneau, Kartikay Khandelwal, Naman Goyal, Vishrav Chaudhary, Guillaume Wenzek, Francisco Guzm{\'a}n, Edouard Grave, Myle Ott, Luke Zettlemoyer, and Veselin Stoyanov. 2020.
\newblock \href {https://doi.org/10.18653/v1/2020.acl-main.747} {Unsupervised cross-lingual representation learning at scale}.
\newblock In \emph{Proceedings of the 58th Annual Meeting of the Association for Computational Linguistics}, pages 8440--8451, Online. Association for Computational Linguistics.

\bibitem[{DataCanary et~al.(2017)DataCanary, hilfialkaff, Jiang, and Meg~Risdal}]{quora-question-pairs}
DataCanary, hilfialkaff, Lili Jiang, and tomtung Meg~Risdal, Nikhil~Dandekar. 2017.
\newblock \href {https://kaggle.com/competitions/quora-question-pairs} {Quora question pairs}.

\bibitem[{Devlin et~al.(2019)Devlin, Chang, Lee, and Toutanova}]{Bert}
Jacob Devlin, Ming-Wei Chang, Kenton Lee, and Kristina Toutanova. 2019.
\newblock \href {https://doi.org/10.18653/v1/N19-1423} {{BERT}: Pre-training of deep bidirectional transformers for language understanding}.
\newblock In \emph{Proceedings of the 2019 Conference of the North {A}merican Chapter of the Association for Computational Linguistics: Human Language Technologies, Volume 1 (Long and Short Papers)}, pages 4171--4186, Minneapolis, Minnesota. Association for Computational Linguistics.

\bibitem[{Dubey et~al.(2024)Dubey, Jauhri, Pandey, Kadian, Al-Dahle, Letman, Mathur, Schelten, Yang, Fan et~al.}]{llama3}
Abhimanyu Dubey, Abhinav Jauhri, Abhinav Pandey, Abhishek Kadian, Ahmad Al-Dahle, Aiesha Letman, Akhil Mathur, Alan Schelten, Amy Yang, Angela Fan, et~al. 2024.
\newblock The llama 3 herd of models.
\newblock \emph{arXiv preprint arXiv:2407.21783}.

\bibitem[{Fan et~al.(2019)Fan, Jernite, Perez, Grangier, Weston, and Auli}]{eli5}
Angela Fan, Yacine Jernite, Ethan Perez, David Grangier, Jason Weston, and Michael Auli. 2019.
\newblock \href {https://doi.org/10.18653/v1/P19-1346} {{ELI}5: Long form question answering}.
\newblock In \emph{Proceedings of the 57th Annual Meeting of the Association for Computational Linguistics}, pages 3558--3567, Florence, Italy. Association for Computational Linguistics.

\bibitem[{Gao et~al.(2023)Gao, Xiong, Gao, Jia, Pan, Bi, Dai, Sun, and Wang}]{gao2023retrieval}
Yunfan Gao, Yun Xiong, Xinyu Gao, Kangxiang Jia, Jinliu Pan, Yuxi Bi, Yi~Dai, Jiawei Sun, and Haofen Wang. 2023.
\newblock Retrieval-augmented generation for large language models: A survey.
\newblock \emph{arXiv preprint arXiv:2312.10997}.

\bibitem[{gbharti(2023)}]{finance-alpaca}
gbharti. 2023.
\newblock \href {https://huggingface.co/datasets/gbharti/finance-alpaca} {finance-alpaca}.
\newblock \emph{Hugging Face}.

\bibitem[{Henderson et~al.(2017)Henderson, Al-Rfou, Strope, Sung, Luk{\'a}cs, Guo, Kumar, Miklos, and Kurzweil}]{loss}
Matthew Henderson, Rami Al-Rfou, Brian Strope, Yun-Hsuan Sung, L{\'a}szl{\'o} Luk{\'a}cs, Ruiqi Guo, Sanjiv Kumar, Balint Miklos, and Ray Kurzweil. 2017.
\newblock Efficient natural language response suggestion for smart reply.
\newblock \emph{arXiv preprint arXiv:1705.00652}.

\bibitem[{Hidey and McKeown(2016)}]{Altlex}
Christopher Hidey and Kathy McKeown. 2016.
\newblock \href {https://doi.org/10.18653/v1/P16-1135} {Identifying causal relations using parallel {W}ikipedia articles}.
\newblock In \emph{Proceedings of the 54th Annual Meeting of the Association for Computational Linguistics (Volume 1: Long Papers)}, pages 1424--1433, Berlin, Germany. Association for Computational Linguistics.

\bibitem[{Hu et~al.(2021)Hu, Shen, Wallis, Allen-Zhu, Li, Wang, Wang, and Chen}]{lora}
Edward~J Hu, Yelong Shen, Phillip Wallis, Zeyuan Allen-Zhu, Yuanzhi Li, Shean Wang, Lu~Wang, and Weizhu Chen. 2021.
\newblock Lora: Low-rank adaptation of large language models.
\newblock \emph{arXiv preprint arXiv:2106.09685}.

\bibitem[{Hu and Lu(2024)}]{hu2024ragrausurvey}
Yucheng Hu and Yuxing Lu. 2024.
\newblock Rag and rau: A survey on retrieval-augmented language model in natural language processing.
\newblock \emph{arXiv preprint arXiv:2404.19543}.

\bibitem[{Jiang et~al.(2023)Jiang, Sablayrolles, Mensch, Bamford, Chaplot, Casas, Bressand, Lengyel, Lample, Saulnier et~al.}]{mistral7b}
Albert~Q Jiang, Alexandre Sablayrolles, Arthur Mensch, Chris Bamford, Devendra~Singh Chaplot, Diego de~las Casas, Florian Bressand, Gianna Lengyel, Guillaume Lample, Lucile Saulnier, et~al. 2023.
\newblock Mistral 7b.
\newblock \emph{arXiv preprint arXiv:2310.06825}.

\bibitem[{Jin et~al.(2019)Jin, Dhingra, Liu, Cohen, and Lu}]{pubmedqa}
Qiao Jin, Bhuwan Dhingra, Zhengping Liu, William Cohen, and Xinghua Lu. 2019.
\newblock \href {https://doi.org/10.18653/v1/D19-1259} {{P}ub{M}ed{QA}: A dataset for biomedical research question answering}.
\newblock In \emph{Proceedings of the 2019 Conference on Empirical Methods in Natural Language Processing and the 9th International Joint Conference on Natural Language Processing (EMNLP-IJCNLP)}, pages 2567--2577, Hong Kong, China. Association for Computational Linguistics.

\bibitem[{Joshi et~al.(2017)Joshi, Choi, Weld, and Zettlemoyer}]{triviaqa}
Mandar Joshi, Eunsol Choi, Daniel Weld, and Luke Zettlemoyer. 2017.
\newblock \href {https://doi.org/10.18653/v1/P17-1147} {{T}rivia{QA}: A large scale distantly supervised challenge dataset for reading comprehension}.
\newblock In \emph{Proceedings of the 55th Annual Meeting of the Association for Computational Linguistics (Volume 1: Long Papers)}, pages 1601--1611, Vancouver, Canada. Association for Computational Linguistics.

\bibitem[{Ju et~al.(2024)Ju, Sun, Du, Yuan, Ren, and Liu}]{llmlayers}
Tianjie Ju, Weiwei Sun, Wei Du, Xinwei Yuan, Zhaochun Ren, and Gongshen Liu. 2024.
\newblock \href {https://aclanthology.org/2024.lrec-main.722} {How large language models encode context knowledge? a layer-wise probing study}.
\newblock In \emph{Proceedings of the 2024 Joint International Conference on Computational Linguistics, Language Resources and Evaluation (LREC-COLING 2024)}, pages 8235--8246, Torino, Italia. ELRA and ICCL.

\bibitem[{Kwiatkowski et~al.(2019)Kwiatkowski, Palomaki, Redfield, Collins, Parikh, Alberti, Epstein, Polosukhin, Devlin, Lee, Toutanova, Jones, Kelcey, Chang, Dai, Uszkoreit, Le, and Petrov}]{nq}
Tom Kwiatkowski, Jennimaria Palomaki, Olivia Redfield, Michael Collins, Ankur Parikh, Chris Alberti, Danielle Epstein, Illia Polosukhin, Jacob Devlin, Kenton Lee, Kristina Toutanova, Llion Jones, Matthew Kelcey, Ming-Wei Chang, Andrew~M. Dai, Jakob Uszkoreit, Quoc Le, and Slav Petrov. 2019.
\newblock \href {https://doi.org/10.1162/tacl_a_00276} {Natural questions: A benchmark for question answering research}.
\newblock \emph{Transactions of the Association for Computational Linguistics}, 7:452--466.

\bibitem[{Lee et~al.(2024)Lee, Roy, Xu, Raiman, Shoeybi, Catanzaro, and Ping}]{NV-Embed}
Chankyu Lee, Rajarshi Roy, Mengyao Xu, Jonathan Raiman, Mohammad Shoeybi, Bryan Catanzaro, and Wei Ping. 2024.
\newblock Nv-embed: Improved techniques for training llms as generalist embedding models.
\newblock \emph{arXiv preprint arXiv:2405.17428}.

\bibitem[{Li et~al.(2023)Li, Zhang, Zhang, Long, Xie, and Zhang}]{qwen2GTE}
Zehan Li, Xin Zhang, Yanzhao Zhang, Dingkun Long, Pengjun Xie, and Meishan Zhang. 2023.
\newblock Towards general text embeddings with multi-stage contrastive learning.
\newblock \emph{arXiv preprint arXiv:2308.03281}.

\bibitem[{Ma et~al.(2024)Ma, Wang, Yang, Wei, and Lin}]{RepLLAMa}
Xueguang Ma, Liang Wang, Nan Yang, Furu Wei, and Jimmy Lin. 2024.
\newblock Fine-tuning llama for multi-stage text retrieval.
\newblock In \emph{Proceedings of the 47th International ACM SIGIR Conference on Research and Development in Information Retrieval}, pages 2421--2425.

\bibitem[{Maia et~al.(2018)Maia, Handschuh, Freitas, Davis, McDermott, Zarrouk, and Balahur}]{fiqa}
Macedo Maia, Siegfried Handschuh, Andr{\'e} Freitas, Brian Davis, Ross McDermott, Manel Zarrouk, and Alexandra Balahur. 2018.
\newblock Www'18 open challenge: financial opinion mining and question answering.
\newblock In \emph{Companion proceedings of the the web conference 2018}, pages 1941--1942.

\bibitem[{Meng et~al.(2024)Meng, Liu, Joty, Xiong, Zhou, and Yavuz}]{SFRAIResearch2024}
Rui Meng, Ye~Liu, Shafiq~Rayhan Joty, Caiming Xiong, Yingbo Zhou, and Semih Yavuz. 2024.
\newblock \href {https://blog.salesforceairesearch.com/sfr-embedded-mistral/} {Sfr-embedding-mistral:enhance text retrieval with transfer learning}.
\newblock Salesforce AI Research Blog.

\bibitem[{Muennighoff et~al.(2024)Muennighoff, Su, Wang, Yang, Wei, Yu, Singh, and Kiela}]{GritLM}
Niklas Muennighoff, Hongjin Su, Liang Wang, Nan Yang, Furu Wei, Tao Yu, Amanpreet Singh, and Douwe Kiela. 2024.
\newblock Generative representational instruction tuning.
\newblock \emph{arXiv preprint arXiv:2402.09906}.

\bibitem[{Muennighoff et~al.(2022)Muennighoff, Tazi, Magne, and Reimers}]{mteb}
Niklas Muennighoff, Nouamane Tazi, Lo{\"\i}c Magne, and Nils Reimers. 2022.
\newblock Mteb: Massive text embedding benchmark.
\newblock \emph{arXiv preprint arXiv:2210.07316}.

\bibitem[{Narayan et~al.(2018)Narayan, Cohen, and Lapata}]{xsum-emnlp}
Shashi Narayan, Shay~B. Cohen, and Mirella Lapata. 2018.
\newblock Don't give me the details, just the summary! {T}opic-aware convolutional neural networks for extreme summarization.
\newblock In \emph{Proceedings of the 2018 Conference on Empirical Methods in Natural Language Processing}, Brussels, Belgium.

\bibitem[{Neelakantan et~al.(2022)Neelakantan, Xu, Puri, Radford, Han, Tworek, Yuan, Tezak, Kim, Hallacy et~al.}]{cpt_text}
Arvind Neelakantan, Tao Xu, Raul Puri, Alec Radford, Jesse~Michael Han, Jerry Tworek, Qiming Yuan, Nikolas Tezak, Jong~Wook Kim, Chris Hallacy, et~al. 2022.
\newblock Text and code embeddings by contrastive pre-training.
\newblock \emph{arXiv preprint arXiv:2201.10005}.

\bibitem[{Oh et~al.(2022)Oh, Kim, Lee, Huang, and Lim}]{lastLayerBert}
Dongsuk Oh, Yejin Kim, Hodong Lee, H.~Howie Huang, and Heuiseok Lim. 2022.
\newblock \href {https://aclanthology.org/2022.coling-1.405} {Don{'}t judge a language model by its last layer: Contrastive learning with layer-wise attention pooling}.
\newblock In \emph{Proceedings of the 29th International Conference on Computational Linguistics}, pages 4585--4592, Gyeongju, Republic of Korea. International Committee on Computational Linguistics.

\bibitem[{Radford et~al.(2018)Radford, Narasimhan, Salimans, Sutskever et~al.}]{casual_attention}
Alec Radford, Karthik Narasimhan, Tim Salimans, Ilya Sutskever, et~al. 2018.
\newblock Improving language understanding by generative pre-training.

\bibitem[{Raffel et~al.(2020)Raffel, Shazeer, Roberts, Lee, Narang, Matena, Zhou, Li, and Liu}]{T5-Encoder}
Colin Raffel, Noam Shazeer, Adam Roberts, Katherine Lee, Sharan Narang, Michael Matena, Yanqi Zhou, Wei Li, and Peter~J Liu. 2020.
\newblock Exploring the limits of transfer learning with a unified text-to-text transformer.
\newblock \emph{Journal of machine learning research}, 21(140):1--67.

\bibitem[{Rajpurkar et~al.(2018)Rajpurkar, Jia, and Liang}]{SQuAD}
Pranav Rajpurkar, Robin Jia, and Percy Liang. 2018.
\newblock Know what you don't know: Unanswerable questions for squad.
\newblock \emph{arXiv preprint arXiv:1806.03822}.

\bibitem[{Reimers(2019)}]{Sentence-BERT}
N~Reimers. 2019.
\newblock Sentence-bert: Sentence embeddings using siamese bert-networks.
\newblock \emph{arXiv preprint arXiv:1908.10084}.

\bibitem[{Springer et~al.(2024)Springer, Kotha, Fried, Neubig, and Raghunathan}]{echoEmbedding}
Jacob~Mitchell Springer, Suhas Kotha, Daniel Fried, Graham Neubig, and Aditi Raghunathan. 2024.
\newblock Repetition improves language model embeddings.
\newblock \emph{arXiv preprint arXiv:2402.15449}.

\bibitem[{Su et~al.(2023)Su, Shi, Kasai, Wang, Hu, Ostendorf, Yih, Smith, Zettlemoyer, and Yu}]{hkuInstructer}
Hongjin Su, Weijia Shi, Jungo Kasai, Yizhong Wang, Yushi Hu, Mari Ostendorf, Wen-tau Yih, Noah~A. Smith, Luke Zettlemoyer, and Tao Yu. 2023.
\newblock \href {https://doi.org/10.18653/v1/2023.findings-acl.71} {One embedder, any task: Instruction-finetuned text embeddings}.
\newblock In \emph{Findings of the Association for Computational Linguistics: ACL 2023}, pages 1102--1121, Toronto, Canada. Association for Computational Linguistics.

\bibitem[{Thorne et~al.(2018)Thorne, Vlachos, Christodoulopoulos, and Mittal}]{fever}
James Thorne, Andreas Vlachos, Christos Christodoulopoulos, and Arpit Mittal. 2018.
\newblock \href {https://doi.org/10.18653/v1/N18-1074} {{FEVER}: a large-scale dataset for fact extraction and {VER}ification}.
\newblock In \emph{Proceedings of the 2018 Conference of the North {A}merican Chapter of the Association for Computational Linguistics: Human Language Technologies, Volume 1 (Long Papers)}, pages 809--819, New Orleans, Louisiana. Association for Computational Linguistics.

\bibitem[{Vaswani et~al.(2017)Vaswani, Shazeer, Parmar, Uszkoreit, Jones, Gomez, Kaiser, and Polosukhin}]{attentionneed}
Ashish Vaswani, Noam Shazeer, Niki Parmar, Jakob Uszkoreit, Llion Jones, Aidan~N. Gomez, Lukasz Kaiser, and Illia Polosukhin. 2017.
\newblock Attention is all you need.

\bibitem[{Wang et~al.(2023)Wang, Yang, Huang, Yang, Majumder, and Wei}]{e5-instruct}
Liang Wang, Nan Yang, Xiaolong Huang, Linjun Yang, Rangan Majumder, and Furu Wei. 2023.
\newblock Improving text embeddings with large language models.
\newblock \emph{arXiv preprint arXiv:2401.00368}.

\bibitem[{Yang et~al.(2024)Yang, Yang, Hui, Zheng, Yu, Zhou, Li, Li, Liu, Huang et~al.}]{qwen2}
An~Yang, Baosong Yang, Binyuan Hui, Bo~Zheng, Bowen Yu, Chang Zhou, Chengpeng Li, Chengyuan Li, Dayiheng Liu, Fei Huang, et~al. 2024.
\newblock Qwen2 technical report.
\newblock \emph{arXiv preprint arXiv:2407.10671}.

\bibitem[{Yang et~al.(2018)Yang, Qi, Zhang, Bengio, Cohen, Salakhutdinov, and Manning}]{hotpotqa}
Zhilin Yang, Peng Qi, Saizheng Zhang, Yoshua Bengio, William Cohen, Ruslan Salakhutdinov, and Christopher~D. Manning. 2018.
\newblock \href {https://doi.org/10.18653/v1/D18-1259} {{H}otpot{QA}: A dataset for diverse, explainable multi-hop question answering}.
\newblock In \emph{Proceedings of the 2018 Conference on Empirical Methods in Natural Language Processing}, pages 2369--2380, Brussels, Belgium. Association for Computational Linguistics.

\bibitem[{Zhang et~al.(2021)Zhang, Ma, Shi, and Lin}]{mr}
Xinyu Zhang, Xueguang Ma, Peng Shi, and Jimmy Lin. 2021.
\newblock \href {https://doi.org/10.18653/v1/2021.mrl-1.12} {Mr. {T}y{D}i: A multi-lingual benchmark for dense retrieval}.
\newblock In \emph{Proceedings of the 1st Workshop on Multilingual Representation Learning}, pages 127--137, Punta Cana, Dominican Republic. Association for Computational Linguistics.

\bibitem[{Zhang et~al.(2022)Zhang, Thakur, Ogundepo, Kamalloo, Alfonso-Hermelo, Li, Liu, Rezagholizadeh, and Lin}]{miracl}
Xinyu Zhang, Nandan Thakur, Odunayo Ogundepo, Ehsan Kamalloo, David Alfonso-Hermelo, Xiaoguang Li, Qun Liu, Mehdi Rezagholizadeh, and Jimmy Lin. 2022.
\newblock Making a miracl: Multilingual information retrieval across a continuum of languages.
\newblock \emph{arXiv preprint arXiv:2210.09984}.

\end{thebibliography}
\bibliographystyle{acl_natbib}
% \input{custom.bbl}
% \bibliography{reference}

% \bibliography{custom}
% \bibliographystyle{acl_natbib}

\appendix

\section{Training and Benchmark Details}
The training datasets for LLM-based embedding models are listed in Table \ref{tab:training_dataset}, and evaluation instructions are listed in Table \ref{tab:eval_dataset}. We use the same instructions as in \citep{e5-instruct} to facilitate easy replication.
\begin{table*}[htbp] 
\small
\centering
\begin{tabular}{p{5cm}rp{6cm}} 
\toprule
    \textbf{Dataset} & \textbf{Examples} & \textbf{Instruction} \\ \hline
STSB \citep{cer-etal-2017-semeval}& $937$ & Retrieve semantically similar text. \\ \midrule
MSMARCO document \citep{MSMARCO}& $73,400$ & Given a web search query, retrieve relevant documents that answer the query. \\ \hline
Quora Duplicates \citep{quora-question-pairs} & $101,762$ & Given a question, retrieve questions that are semantically equivalent to the given question. \\ \hline
MSMARCO passage \citep{MSMARCO}& $249,592$ & Given a web search query, retrieve relevant passages that answer the query. \\ \hline
NQ \citep{nq} & $100,231$ & Given a question, retrieve Wikipedia passages that answer the question. \\ \hline
SQuAD \citep{SQuAD} & $87,599$ & Retrieve Wikipedia passages that answer the question. \\ \hline
TriviaQA \citep{triviaqa} & $73,346$ & Retrieve Wikipedia passages that answer the question. \\ \hline
AllNLI \citep{allnil} & $277,230$ & Given a premise, retrieve a hypothesis that is entailed by the premise. \\ \hline
finance-alpaca \citep{finance-alpaca} & $35,038$ & Given a finance question, retrieve passages that answer the question. \\ \hline
FiQA \citep{fiqa} & $7,203$ & Given a finance question, retrieve passages that answer the question. \\ \hline
MIRACL \citep{miracl} & $32,561$ & Given a question, retrieve passages that answer the question. \\ \hline
xsum \citep{xsum-emnlp} & $24,626$ & Given a document, retrieve semantically similar summaries. \\ \hline
Mr. TYDI \citep{mr} & $48,715$ & Given a question, retrieve Wikipedia passages that answer the question. \\ \hline
Altlex \citep{Altlex} & $54,674$ & Retrieve semantically similar text. \\ \hline
HotpotQA \citep{hotpotqa}& $90,447$ & Given a multi-hop question, retrieve documents that can help answer the question. \\ \hline
ELI5 \citep{eli5}  & $32,547$ & Provided a user question, retrieve the highest voted answers on Reddit ELI5 forum. \\ \hline
FEVER \citep{fever} & $101,578$ & Given a claim, retrieve documents that support or refute the claim. \\ \hline
PubMedQA \citep{pubmedqa} & $500$ & Given a biomedical question, retrieve information that answers the question. \\ \bottomrule
\end{tabular}
\caption{Training Dataset Overview and Instructions} 
\label{tab:training_dataset} 
\end{table*}

\begin{table*}[htb]
\scriptsize
\centering
\begin{tabular}{p{5cm}p{8cm}}
\toprule
\textbf{Dataset} & \textbf{Instruction} \\
\toprule
NFCorpus & Given a question, retrieve relevant documents that best answer the question. \\
\midrule
ArguAna & Given a claim, find documents that refute the claim. \\
\hline
ClimateFEVER & Given a claim about climate change, retrieve documents that support or refute the claim. \\
\hline
DBPedia & Given a query, retrieve relevant entity descriptions from DBPedia. \\
\hline
FEVER & Given a claim, retrieve documents that support or refute the claim. \\
\hline
FiQA2018 & Given a financial question, retrieve user replies that best answer the question. \\
\hline
HotpotQA & Given a multi-hop question, retrieve documents that can help answer the question. \\
\hline
MSMARCO & Given a web search query, retrieve relevant passages that answer the query. \\
\hline
NQ & Given a question, retrieve Wikipedia passages that answer the question. \\
\hline
QuoraRetrieval & Given a question, retrieve questions that are semantically equivalent to the given question. \\
\hline
SCIDOCS & Given a scientific paper title, retrieve paper abstracts that are cited by the given paper. \\
\hline
SciFact & Given a scientific claim, retrieve documents that support or refute the claim. \\
\hline
Touche2020 & Given a question, retrieve detailed and persuasive arguments that answer the question. \\
\hline
TRECCOVID & Given a query, retrieve documents that answer the query. \\
\hline
FinanceBench & Given a financial question, retrieve user replies that best answer the question. \\
\hline
Company2Industry & Given a company name, retrieve the related industry. \\
\hline
BIOSSES & Retrieve semantically similar text. \\
\hline
SICK-R & Retrieve semantically similar text. \\
\hline
STS12 & Retrieve semantically similar text. \\
\hline
STS13 & Retrieve semantically similar text. \\
\hline
STS14 & Retrieve semantically similar text. \\
\hline
STS15 & Retrieve semantically similar text. \\
\hline
STS16 & Retrieve semantically similar text. \\
\hline
STSBenchmark & Retrieve semantically similar text. \\
\hline
AskUbuntuDupQuestions & Retrieve duplicate questions from AskUbuntu forum. \\
\hline
MindSmallReranking & Retrieve relevant news articles based on user browsing history. \\
\hline
SciDocsRR & Given a title of a scientific paper, retrieve the titles of other relevant papers. \\
\hline
StackOverflowDupQuestions & Retrieve duplicate questions from StackOverflow forum. \\
\hline
AmazonPolarityClassification & Classify Amazon reviews into positive or negative sentiment. \\
\hline
ToxicConversationsClassification & Classify the given comments as either toxic or not toxic. \\
\hline
Banking77Classification & Given an online banking query, find the corresponding intents. \\
\hline
EmotionClassification & Classify the emotion expressed in the given Twitter message into one of the six emotions: anger, fear, joy, love, sadness, and surprise. \\
\hline
ImdbClassification & Classify the sentiment expressed in the given movie review text from the IMDB dataset. \\
\hline
TweetSentimentExtractionClassification & Classify the sentiment of a given tweet as either positive, negative, or neutral. \\
\hline
SummEval & Given a news summary, retrieve other semantically similar summaries. \\
\hline
TwentyNewsgroupsClustering & Identify the topic or theme of the given news articles. \\
\hline
ArxivClusteringP2P & Identify the main and secondary category of Arxiv papers based on the titles and abstracts. \\
\hline
ArxivClusteringS2S & Identify the main and secondary category of Arxiv papers based on the titles. \\
\hline
BiorxivClusteringP2P.v2 & Identify the main category of Biorxiv papers based on the titles and abstracts. \\
\hline
BiorxivClusteringS2S.v2 & Identify the main category of Biorxiv papers based on the titles. \\
\hline
MedrxivClusteringP2P.v2 & Identify the main category of Medrxiv papers based on the titles and abstracts. \\
\hline
MedrxivClusteringS2S.v2 & Identify the main category of Medrxiv papers based on the titles. \\
\hline
RedditClustering.v2 & Identify the topic or theme of Reddit posts based on the titles. \\
\hline
RedditClusteringP2P.v2 & Identify the topic or theme of Reddit posts based on the titles and posts. \\
\hline
StackExchangeClustering.v2 & Identify the topic or theme of StackExchange posts based on the titles. \\
\hline
StackExchangeClusteringP2P.v2 & Identify the topic or theme of StackExchange posts based on the given paragraphs. \\
\hline
TwentyNewsgroupsClustering.v2 & Identify the topic or theme of the given news articles. \\
\hline
TwitterURLCorpus & Retrieve tweets that are semantically similar to the given tweet. \\
\hline
SprintDuplicateQuestions & Retrieve duplicate questions from Sprint forum. \\
\hline
TwitterSemEval2015 & Retrieve tweets that are semantically similar to the given tweet. \\
\bottomrule
\end{tabular}
\caption{Evaluation Instructions for MTEB Benchmark} 
\label{tab:eval_dataset} 
\end{table*}

\label{sec:implementation}

\section{Detailed MTEB Evaluation Results}
% \label{sec:mteb_full}
\begin{table}[htb]
\centering
 \resizebox{\textwidth}{!}{
\begin{tabular}{ccccccccccc}
\toprule
Combination & BaseModel & Avg. & BIOSSES & SICK-R & STS12 & STS13 & STS14 & STS15 & STS16 & STSBenchmark \\
\midrule
Model 1 & Mistral-7B   & 0.8302 & 0.8530&	0.8255&	0.7147&	0.8464&	0.8055&	0.8777&	0.8529&	0.8659 \\

Model 2 & Mistral-7B   &0.8431& 0.8699&0.8309	&0.7424&0.8553&0.8227&0.8858&0.8593&0.8786\\

Model 3 & Mistral-7B   &0.8420 & 0.8641	&0.8316	&0.7410	&0.8589	&0.8195	&0.8839& 0.8596	&0.8777 \\

Model 4 & Mistral-7B  &0.8397& 0.8650&	0.8392	&0.7368	&0.8447	&0.8200	&0.8811	&0.8547	&0.8763\\

Model 5 & Mistral-7B  &0.8468 & 0.8808&	0.8389	&0.7441	&0.8572&	0.8322&0.8844	&0.8585&	0.8782\\ 
\bottomrule
\end{tabular}
}
\caption{MTEB results on STS tasks.}
\label{tab:STS} 
\end{table}

\begin{table}[htbp]
\centering
 \resizebox{\textwidth}{!}{
\begin{tabular}{ccccccccccc}
\toprule
Combination & Model & Avg. & ArguAna & ClimateFEVER & DBPedia & FEVER & FiQA2018 & HotpotQA & MSMARCO \\
\midrule
Model 1 & Mistral-7B  & 0.5394 & 0.4863 & 0.3814 & 0.4828 & 0.9076 & 0.5262 & 0.6567 & 0.4158 \\
Model 2 & Mistral-7B  & 0.5529 & 0.5408 & 0.4009 & 0.4579 & 0.9109 & 0.5582 & 0.6493 & 0.4175 \\
Model 3 & Mistral-7B   & 0.5496 & 0.5052 & 0.4206 & 0.4524 & 0.9119 & 0.5553 & 0.6689 & 0.4198 \\
Model 4 & Mistral-7B  & 0.5607 & 0.5489 & 0.3772 & 0.4502 & 0.9177 & 0.5888 & 0.6891 & 0.4319 \\
Model 5 & Mistral-7B  & 0.5620 & 0.5551 & 0.4024 & 0.4432 & 0.9150 & 0.5747 & 0.6852 & 0.4261 \\
\bottomrule
\end{tabular}
}
\caption{MTEB results on Retrieval tasks: Part 1.}
\label{tab:first_half}
\end{table}

\begin{table}[htbp]
\centering
 \resizebox{\textwidth}{!}{
\setlength{\tabcolsep}{3pt} % Reduce space between columns
\begin{tabular}{ccccccccc}
\toprule
Combination & Model & NFCorpus & NQ & QuoraRetrieval & SCIDOCS & SciFact & Touche2020 & TRECCOVID \\
\midrule
Model 1 & Mistral-7B & 0.3609 & 0.6433 & 0.8894 & 0.1903 & 0.7582 & 0.2359 & 0.6175 \\
Model 2 & Mistral-7B   & 0.3748 & 0.6413 & 0.8910 & 0.1881 & 0.7332 & 0.2381 & 0.6938 \\
Model 3 & Mistral-7B  & 0.3699 & 0.6429 & 0.8900 & 0.1889 & 0.7474 & 0.2344 & 0.7322 \\
Model 4 & Mistral-7B  & 0.3901 & 0.6543 & 0.8931 & 0.2023 & 0.7418 & 0.2348 & 0.7291 \\
Model 5 & Mistral-7B   & 0.3808 & 0.6636 & 0.8923 & 0.1980 & 0.7403 & 0.2414 & 0.7505 \\
\bottomrule
\end{tabular}
}
\caption{MTEB results on Retrieval tasks: Part 2.}
\label{tab:second_half}
\end{table}

\begin{table}[htbp]
\centering
 \resizebox{\textwidth}{!}{
\begin{tabular}{cccccccc}
\toprule
Combination & Model &  Banking77 &   Emotion &   TweetSentiment & Amazon Polarity & Toxic Conversations & Imdb \\
\midrule
Model 1 & Mistral-7B  &  0.8269 & 0.4871 & 0.6044 & 0.9161 & 0.6342 & 0.8777  \\
Model 2 & Mistral-7B   &0.8319 & 0.4610 & 0.6048 & 0.9161 & 0.6340 & 0.8777 \\
Model 3 & Mistral-7B   & 0.8353 & 0.4875 & 0.6029 & 0.9108 & 0.6249 & 0.8650  \\
Model 4 & Mistral-7B   &0.8345 & 0.4638 & 0.5773 & 0.8647 & 0.5973 & 0.7189 \\
Model 5 & Mistral-7B  & 0.8304 & 0.5037 & 0.6067 & 0.9069 & 0.6275 & 0.7853 \\
\bottomrule
\end{tabular}
}
\caption{MTEB results on Classification tasks.}
\label{tab:class}
\end{table}

\begin{table}[htbp]
\centering
 \resizebox{\textwidth}{!}{
\begin{tabular}{cccccccc}
\toprule
Combination & Model & ArxivClusteringP2P & ArxivClusteringS2S & BiorxivClusteringP2P & BiorxivClusteringS2S & MedrxivClusteringP2P & MedrxivClusteringS2S  \\
\midrule
Model 1 & Mistral-7B  & 0.4896 & 0.4462 & 0.3840 & 0.3686 & 0.3334 & 0.3248 \\
Model 2 & Mistral-7B & 0.4842 & 0.4469 & 0.3838 & 0.3756 & 0.3356 & 0.3339  \\
Model 3 & Mistral-7B  & 0.4844 & 0.4559 & 0.3823 & 0.3765 & 0.3398 & 0.3436  \\
Model 4 & Mistral-7B    & 0.4419 & 0.4212 & 0.3457 & 0.3289 & 0.3309 & 0.3068  \\
Model 5 & Mistral-7B  & 0.4653 & 0.4314 & 0.3549 & 0.3534 & 0.3366 & 0.3291 \\
\bottomrule
\end{tabular}
}
\caption{MTEB results on Clustering tasks: Part 1.}
\label{tab:custering_1}
\end{table}

\begin{table}[htbp]
\centering
\begin{adjustbox}{width=\textwidth,center}
\begin{tabular}{ccccccc}
\toprule
Combination & Model & RedditClusteringP2P & StackExchangeClustering & StackExchangeClusteringP2P & TwentyNewsgroupsClustering & RedditClustering \\
\midrule
Model 1 & Mistral-7B  & 0.6408 & 0.5569 & 0.3982 & 0.4931 & 0.5182\\
Model 2 & Mistral-7B   & 0.6389 & 0.4972 & 0.4155 & 0.4461 & 0.5114 \\
Model 3 & Mistral-7B   & 0.6381 & 0.5003 & 0.4099 & 0.4718 & 0.5324\\
Model 4 & Mistral-7B    & 0.5782 & 0.3944 & 0.3861 & 0.4329& 0.4441 \\
Model 5 & Mistral-7B & 0.6190 & 0.4516 & 0.3966 & 0.4455 & 0.4997 \\
\bottomrule
\end{tabular}
\end{adjustbox}
\caption{MTEB results on Clustering tasks: Part 2.}
\label{tab:custering_2}
\end{table}

% \begin{table}[htbp]
% \centering
% \begin{adjustbox}{width=\textwidth,center}
% \tiny 
% \begin{tabular}{lcccc}
% \toprule
% Combination & Model &   Avg.&   BTwitterURLCorpus	& SprintDuplicate  Questions& TwitterSemEval2015 \\
% \midrule
% Model 1 & Mistral-7B  & 0.8605 &0.8645	&0.9491	&0.7679  \\
% Model 3 & Mistral-7B  & 0.8627&0.8644	&0.9571&	0.7666  \\
% Model 2 & Mistral-7B  & 0.8639& 0.8660&	0.9553	&0.7704 \\
% Model 5 & Mistral-7B  & 0.8746&0.8675&	0.9653&	0.7910\\
% Model 4 & Mistral-7B  & 0.8707& 0.8617&	0.9643	&0.7860 \\
% \bottomrule
% \end{tabular}
% \end{adjustbox}
% \caption{MTEB results on Pair Classification tasks.}
% \label{tab:pair_class}
% \end{table}

% \begin{table}[htbp]
% \centering
% \begin{adjustbox}{width=\textwidth,center}
% \tiny 
% \begin{tabular}{lccccc}
% \toprule
% Combination & Model &   Avg.&   AskUbuntuDupQuestions	& MindSmallReranking&  SciDocsRR& StackOverflowDupQuestions \\
% \midrule
% Model 1 & Mistral-7B   &0.5737 &0.6480 &0.2859 &0.8341 &0.5267  \\
% Model 3 & Mistral-7B  &0.5787 &0.6482 &0.2930 &0.8383 &0.5353   \\
% Model 2 & Mistral-7B   &0.5720 &0.6459 &0.2855 &0.8302 &0.5263\\
% Model 5 & Mistral-7B   &0.5812 &0.6645 &0.2983 &0.8248 &0.5373 \\
% Model 4 & Mistral-7B  &0.5829 &0.6667 &0.3139 &0.8120 &0.5388  \\
% \bottomrule
% \end{tabular}
% \end{adjustbox}
% \caption{MTEB results on Reranking tasks.}
% \label{tab:rerank}
% \end{table}
The detailed MTEB \citep{mteb} evaluation results for each task 
 across different embedding models are illustrated in the \cref{tab:STS,tab:first_half,tab:second_half,tab:class,tab:custering_1,tab:custering_2}. The reported metrics represent the main scores for each task as defined by MTEB. Specifically, we report the Spearman correlation of cosine similarity for the Semantic Textual Similarity (STS) task. We utilize the Normalized Discounted Cumulative Gain (NDCG) at rank 10 for the Retrieval task. For the Classification task, we report Accuracy. Lastly, for the Clustering task, the V-measure metric is used.

\end{document}